\documentclass{article}

\usepackage{arxiv}
\usepackage[utf8]{inputenc} % allow utf-8 input
\usepackage[T1]{fontenc}    % use 8-bit T1 fonts
\usepackage{hyperref}       % hyperlinks
\usepackage{url}            % simple URL typesetting
\usepackage{booktabs}       % professional-quality tables
\usepackage{amsfonts}       % blackboard math symbols
\usepackage{nicefrac}       % compact symbols for 1/2, etc.
\usepackage{microtype}      % microtypography
\usepackage{lipsum}		% Can be removed after putting your text content
\usepackage{graphicx}
\usepackage{natbib}
\usepackage{doi}
\usepackage{amsmath}
\usepackage{tikz}
\usepackage{pgfplots}
\usepackage{subcaption}
\usepackage{float}

\hypersetup{colorlinks,linkcolor={blue},citecolor={blue},urlcolor={red}}  

\title{Detecting Deepfake Talking Heads from \\ Facial Biometric Anomalies}
\date{} 					% Or removing it

\author{Justin D. Norman and Hany Farid\\
\\
University of California, Berkeley\\
{\tt\small justin.norman@berkeley.edu and hfarid@berkeley.edu}
}

\renewcommand{\headeright}{}
\renewcommand{\undertitle}{}
\renewcommand{\shorttitle}{Detecting Deepfake Talking Heads}

\hypersetup{
}

\begin{document}
\maketitle

\begin{abstract}
The combination of highly realistic voice cloning, along with visually compelling avatar, face-swap, or lip-sync deepfake video generation, makes it relatively easy to create a video of anyone saying anything. Today, such deepfake impersonations are often used to power frauds, scams, and political disinformation. We propose a novel forensic machine learning technique for the detection of deepfake video impersonations that leverages unnatural patterns in facial biometrics. We evaluate this technique across a large dataset of deepfake techniques and impersonations, as well as assess its reliability to video laundering and its generalization to previously unseen video deepfake generators.
\end{abstract}

\section{Introduction}

\begin{figure*}[t]
    \centerline{\includegraphics[width=17cm]{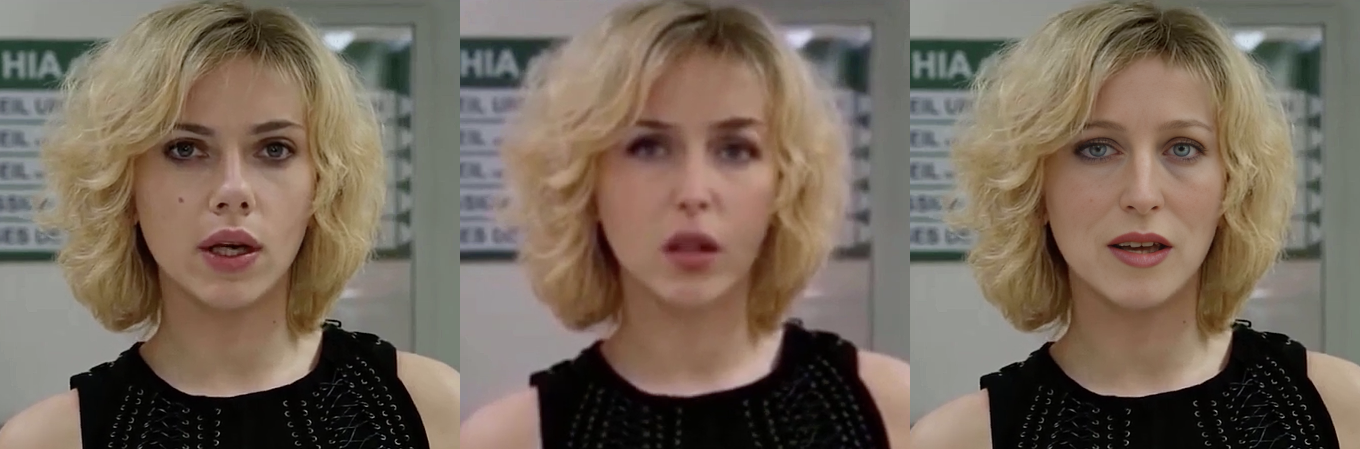}}
\caption{A face-swap deepfake in which the identity in an original video frame (left) is replaced with another identity using deepfake technology from 2019 (middle) and 2025 (right).}
% right image was created using swapface and the identity is Gillian Anderson. 
% center image is taken from an old presentation I gave and I don't know who that person is.
\label{fig:deepfake-2019-2025}
\end{figure*}

It has been some eight years since the term deepfake splashed onto the screen. With its roots in the creation of non-consensual intimate imagery (NCII), in which one person's likeness is replaced with another in sexually explicit material, deepfake technology (re-branded as generative AI starting in 2022) has continued its ballistic trajectory in terms of quality and accessibility. Shown in Figure~\ref{fig:deepfake-2019-2025}, for example, is a face-swap deepfake in which the identity in an original video frame (left) is replaced with another identity using deepfake technology from 2019 (middle) and 2025 (right). Whereas face-swap deepfakes in 2019 were low-resolution and temporally glitchy, today's face-swap deepfakes are nearly flawless. At the same time, the software to create these deepfakes has become easier to use and more ubiquitous.

Deepfakes continue to be weaponized in the creation of NCII, and more recently, in the creation of child sexual abuse material (CSAM). Deepfakes are also used to bolster dangerous lies, conspiracies, disinformation campaigns, and a wide range of small- to large-scale financial frauds. The rise of deepfakes has brought more attention to the development of forensic techniques to detect all forms of image, audio, and video deepfakes~\cite{farid2022creating}.

The creation of video deepfakes fall into two broad categories: impersonation and text-to-video. Here we will focus only on impersonation deepfakes. Three of the most popular techniques are face-swap, lip-sync and avatar deepfakes. In a face-swap deepfake, a person's face in an original video is replaced with another~\cite{nirkin2019fsgan}; in a lip-sync deepfake, a person's mouth region is modified to be consistent with a new voice track~\cite{suwajanakorn2017}; and in an avatar deepfake the head movements and facial expressions of one person are used to animate a single image of another person~\cite{thies2016face2face}.

The detection of video deepfakes falls into one of three broad categories: data-, artifact-, and identity-based.  

Data-based techniques are arguably the most dominant approaches in which a variety of tools from machine learning are used to extract artifacts that arise from the synthesis process that (typically) are not visually salient. In these supervised-learning approaches, a machine-learning model is trained to distinguish real from synthesized content. Depending on the initial representation, network architecture, and objective function being optimized, these networks can learn different features to distinguish the real from the synthesized~\cite{zhou2017twostream,afchar2018mesonet,amerini2019deepfake,zhao2021multi,patil2024genconvit,tan2024frequency,liu2024lips}. The advantage of these data-based approaches is they are able to learn detailed and subtle artifacts resulting from video synthesis. The disadvantage is these approaches typically require large amounts of labeled training data and often struggle to generalize to new synthetic content not seen during training. These techniques can also be vulnerable to adversarial attacks and to simple laundering attacks where the synthetic media is trans-coded and resized~\cite{barni2018adversarial}.

Artifact-based techniques are designed to detect specific spatio-temporal artifacts that emerge during the deepfake generation including, for example, unnatural eye blinks~\cite{li2018ictu}, splicing artifacts~\cite{li2019xray}, anomalous head poses~\cite{yang2019}, inconsistencies between the dynamics of the mouth shape and a spoken word~\cite{agarwal2020phoneme}, and physiological signals~\cite{ciftci2020hearts}. The advantage of these approaches is that they tend to be less data intensive and more resilient to laundering. The disadvantage is that as new deepfake generators emerge, they may overcome previous limitations so these techniques can somewhat of a moving target.

Lastly, identity-based techniques are based on the assumption that as a person speaks, they have distinct facial expressions and movements that can be used to distinguish them from others, impersonators, and deepfakes
~\cite{agarwal2019protecting,dong2020identity,bohacek2022protecting,cozzolino2023audio}. The advantage of these approaches is that they exploit a fundamental limitation of deepfakes -- the person talking is not who it purports to be. The disadvantage is that a customized model is required for each individual making this approach unlikely to scale.

Our work is most similar to identity-based approaches, but unlike previous techniques -- which require building a model for a specific individual -- our technique is identity agnostic, leveraging facial biometric anomalies with no need for an external reference. In particular, we show that, over the length of a video, the variation in the facial appearance of a face-swap deepfake ranges from being anomalously large to anomalously small relative to a natural face, Figure~\ref{fig:pairwise-biometrics}. We posit that large variations in the facial biometric identity are due to structural errors in mapping one identity onto another, while unusually small variations are due complex facial movements and expressions not being well captured.

We describe the methodology behind this approach, followed by extensive validation on a large dataset with three different face-swap deepfakes, a evaluation of robustness to video resolution and compression quality, and in-the-wild deepfakes.

\section{Methods}
\label{sec:methods}

\subsection{Dataset}
\label{subsec:dataset}

The DeepSpeak dataset (v1.0)~\cite{barrington2024deepspeak} contains $21.2$ hours of authentic video of $219$ distinct identities speaking in front of their web camera. This dataset also contains $26.8$ hours of a combination of face-swap and lip-sync deepfakes. The face-swap deepfake videos were partitioned into a training, testing, and validation dataset consisting of $12.0$/$7.4$, $2.4$/$1.0$, and $2.4$/$1.0$ hours of authentic/deepfake video and $116$, $25$, and $25$ identities, with no overlap in identities across these three subsets. The lipsync deepfake videos consist of $2.4$/$2.4$, $0.51$/$0.51$, and $0.55$/$0.55$ hours of training, testing and validation authentic/deepfake video and $66$, $16$, and $14$ identities, with no overlap in identities.

The original DeepSpeak dataset partitions the authentic and deepfake videos of each identity into multiple short individual videos. For each identity, and for each authentic/deepfake type, we concatenated all individual videos for each identity. At $30$ frames per second these combined authentic videos ranged in length from $3$ to $403$ seconds, with $50$ to $338$ seconds of face-swap deepfakes, and $165$ seconds of lip-sync deepfakes per identity.

Each video in this dataset was pre-processed as follows: (1) each video frame was extracted and the location of the face localized using Dlib~\cite{king2009dlib}; (2) The head pose (pitch, yaw, and roll) was estimated using Dlib and OpenCV's solvePnP; and (3) Any non-facial body parts or objects obscuring the face were detected using MediaPipe's Holistic~\cite{mediapipe}. A video frame was discarded if no face was detected,  if the head pitch, yaw or roll angle was greater than $25$, $20$, or $20$ degrees, or if the detected face detected was partially obscured.

\begin{figure*}[t]
    \centering
    \includegraphics[width=4.0cm]{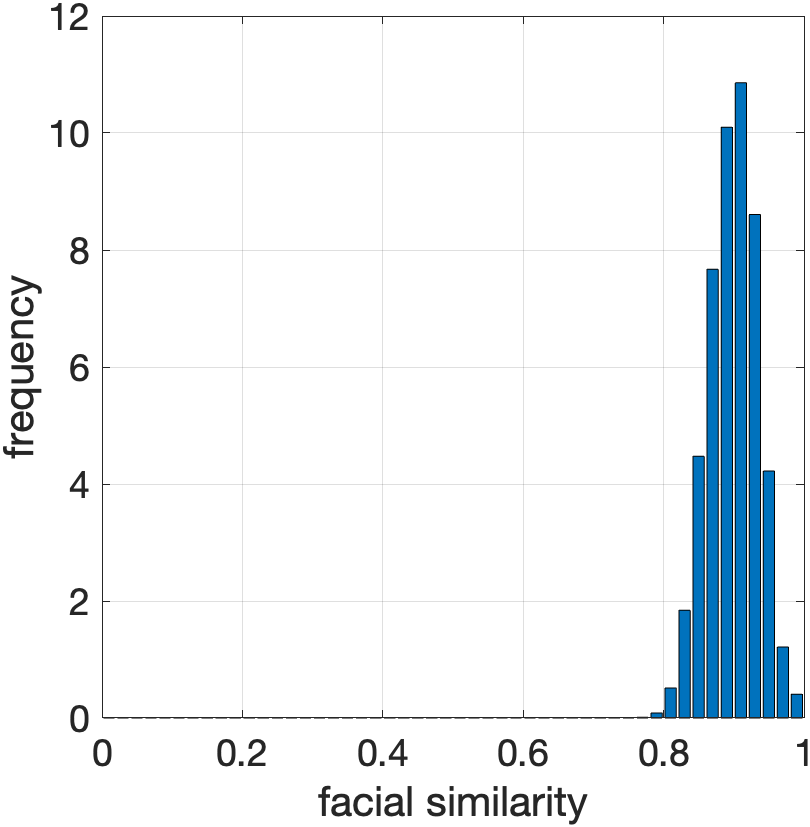}\hfill%
    \includegraphics[width=4.0cm]{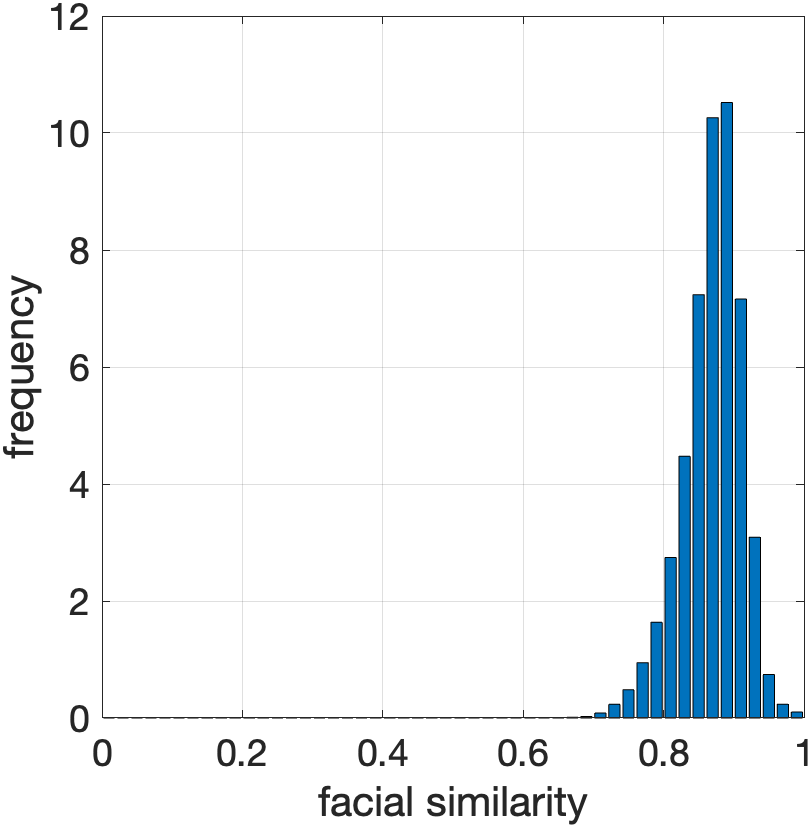}\hfill%
    \includegraphics[width=4.0cm]{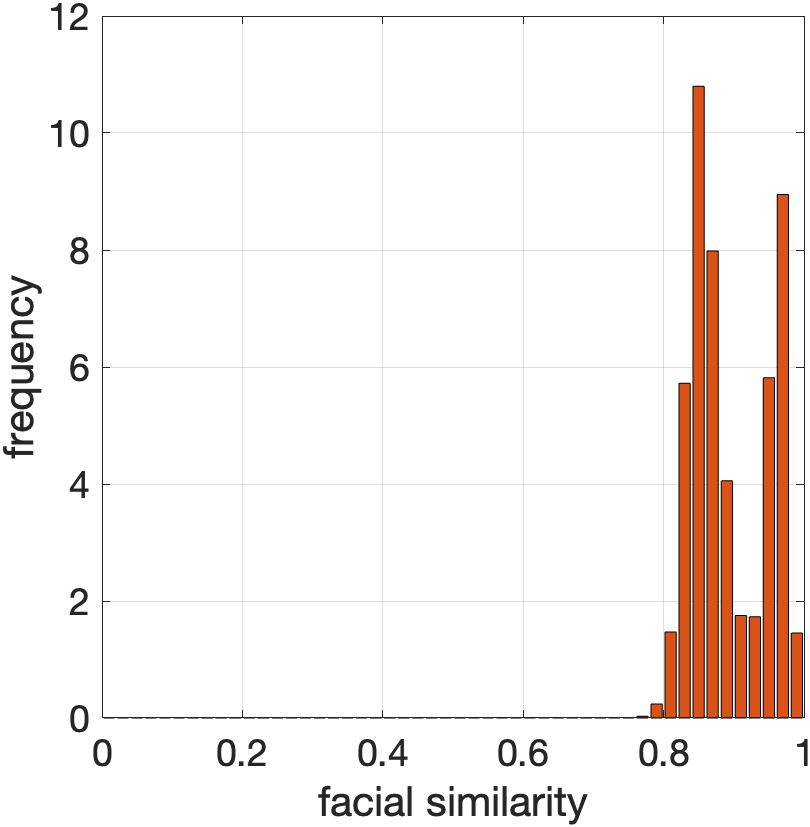}\hfill%
    \includegraphics[width=4.0cm]{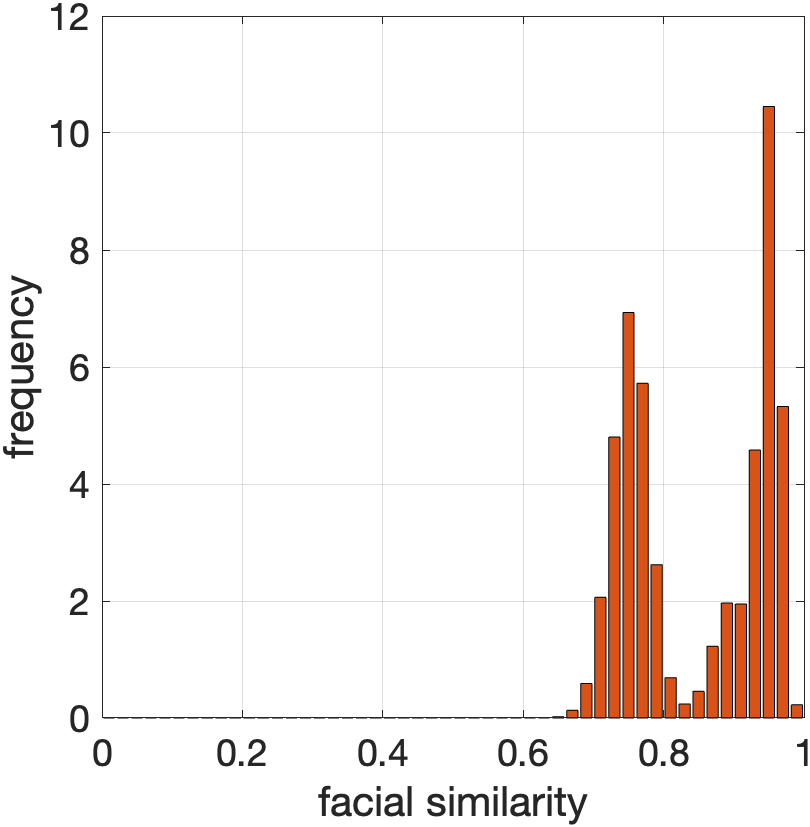}
    \caption{Distribution of pairwise biometric similarity (on a scale of $[-1,1]$) between the extracted face in all video frames for two authentic identities (left, blue) and two face-swap deepfakes (right, red).}
    \label{fig:pairwise-biometrics}
\end{figure*}

\subsection{Biometrics}

The basis for many of the state-of-the-art recognition systems is the CNN-based architecture ArcFace~\cite{deng2019arcface}. For a localized face in an image or video frame, ArcFace extracts a $512$-D representation that has been shown to be an effective biometric marker~\cite{k2020deep}, despite some limitations~\cite{komkov2021advhat, deng2020sub}.

We quantify the similarity between two, unit-length face biometric vectors $\vec{v}_1$ and $\vec{v}_2$ as the cosine similarity $\cos^{-1}(\vec{v}_1 \cdot \vec{v}_2)$. This metric yields a similarity in the range of $-1$ (least similar) to $1$ (most similar). Shown in Figure~\ref{fig:pairwise-biometrics} is a distribution of cosine similarity between all-pairs of faces in two authentic and two face-swap deepfake videos.

\subsection{Feature Engineering}
\label{subsec:feature-engineering}

We seek to quantify the differences in the authentic/deepfake distributions like those shown in Figure~\ref{fig:pairwise-biometrics}. We considered a variety of approaches from fitting parametric or non-parametric density functions to quantifying the distributions from their discrete histogram counts or statistical moments. We settled on the latter because in exploratory analyses, it afforded better overall discriminatory power. 

In particular, given $N$ video frames with a valid face (see Section~\ref{subsec:dataset}), we extract the following discrete features from the $N(N-1)/2$ pairwise face biometric similarities:
\begin{enumerate}
    \item The basic statistical moments are quantified with the: mean, variance, skewness, kurtosis, and the 25-th, 50-th, 75-th quantiles.
    \item The variance-to-mean ratio (or index of dispersion) quantifies the degree of dispersion or clustering within a distribution, while the kurtosis-to-variance ratio captures the shape of the distribution in relation to its overall spread.
\end{enumerate} 
Taken together, these features yield a $9$D feature vector which, as described next, are used to distinguish authentic from deepfake videos.

\begin{table*}[t]
\centering
\caption{Performance metrics (Accuracy, Precision, Recall, and F1) across different video lengths (number of frames), training configurations, and deepfake videos.}
\vspace{0.25cm}
\small  % Uncomment this line
\begin{tabular}{c|cc|c|cc|cccc}
\toprule
& \textbf{Video} & \textbf{Train} & \textbf{Eval} & \textbf{Train} & \textbf{Eval} & \multicolumn{4}{c}{\textbf{Evaluation (\%)}} \\
& \textbf{Length} & \textbf{Model} & \textbf{Model} & \textbf{Size} & \textbf{Size} & \textbf{Acc} & \textbf{Prec} & \textbf{Recall} & \textbf{F1}\\
\midrule
1 & 5000 & FaceFusion & FaceFusion & 66 & 14 & 96.6 & 96.8 & 96.6 & 96.5 \\
2 & 5000 & FaceFusion & FaceFusion + GAN & 66 & 14 & 96.7 & 97.5 & 96.6 & 96.7 \\
3 & 5000 & FaceFusion & FaceFusion Live & 66 & 14 & 82.4 & 88.3 & 81.1 & 82.3 \\
\midrule
4 & 5000 & FaceFusion + GAN & FaceFusion & 66 & 14 & 96.7 & 96.7 & 96.7 & 96.7 \\
5 & 5000 & FaceFusion + GAN & FaceFusion + GAN & 66 & 14 & 96.5 & 96.9 & 96.5 & 96.7 \\
6 & 5000 & FaceFusion + GAN & FaceFusion Live & 66 & 14 & 85.6 & 89.5 & 85.0 & 85.3 \\
\midrule
7 & 5000 & FaceFusion Live & FaceFusion & 66 & 14 & 86.2 & 89.6 & 85.6 & 86.0 \\
8 & 5000 & FaceFusion Live & FaceFusion + GAN & 66 & 14 & 84.7 & 89.0 & 83.7 & 84.7 \\
9 & 5000 & FaceFusion Live & FaceFusion Live & 66 & 14 & 94.7 & 96.1 & 94.4 & 94.3 \\
\midrule
10 & 5000 & Combined & FaceFusion & 66 & 14 & 98.3 & 98.6 & 98.3 & 98.3 \\
11 & 5000 & Combined & FaceFusion + GAN & 66 & 14 & 98.3 & 98.6 & 98.3 & 98.3 \\
12 & 5000 & Combined & FaceFusion Live & 66 & 14 & 85.6 & 89.5 & 85.0 & 85.3 \\
\midrule
13 & Unbounded & Combined & FaceFusion & 152 & 26 & 94.8 & 95.3 & 94.8 & 94.8 \\
14 & Unbounded & Combined & FaceFusion + GAN & 152 & 26 & 94.8 & 95.3 & 94.8 & 94.8 \\
15 & Unbounded & Combined & FaceFusion Live & 152 & 26 & 94.3 & 94.3 & 94.3 & 94.3 \\
\midrule
16 & 4000 & FaceFusion & FaceFusion & 84  & 18 & 96.2 & 96.3 & 96.2 & 96.2 \\
17 & 3000 & FaceFusion & FaceFusion & 93  & 20 & 96.1 & 96.3 & 96.1 & 96.1 \\
18 & 2000 & FaceFusion & FaceFusion & 98  & 21 & 94.7 & 95.7 & 94.7 & 95.0 \\
19 & 1000 & FaceFusion & FaceFusion & 98  & 21 & 92.2 & 92.9 & 92.2 & 92.2 \\
20 & 500 & FaceFusion  & FaceFusion & 103 & 22 & 89.5 & 91.9 & 89.1 & 89.3 \\
\midrule
21 & 4000 & FaceFusion + GAN & FaceFusion + GAN & 84  & 18 & 96.0 & 96.1 & 96.0 & 96.1 \\
22 & 3000 & FaceFusion + GAN & FaceFusion + GAN & 93  & 20 & 92.4 & 92.5 & 92.4 & 92.4 \\
23 & 2000 & FaceFusion + GAN & FaceFusion + GAN & 93  & 20 & 92.7 & 92.9 & 92.7 & 92.7 \\
24 & 1000 & FaceFusion + GAN & FaceFusion + GAN & 103 & 22 & 91.1 & 91.7 & 90.5 & 89.8 \\
25 & 500 & FaceFusion + GAN  & FaceFusion + GAN & 121 & 26 & 85.2 & 88.9 & 92.9 & 81.8 \\
\midrule
26 & 4000 & FaceFusion Live & FaceFusion Live & 97 & 21  & 97.3 & 97.4 & 97.3 & 96.1 \\
27 & 3000 & FaceFusion Live & FaceFusion Live & 117 & 25 & 97.0 & 97.1 & 97.0 & 97.0 \\
28 & 2000 & FaceFusion Live & FaceFusion Live & 117 & 25 & 95.8 & 95.8 & 95.8 & 95.8 \\
29 & 1000 & FaceFusion Live & FaceFusion Live & 117 & 25 & 94.0 & 94.0 & 93.9 & 94.0 \\
30 & 500 & FaceFusion Live  & FaceFusion Live & 121 & 26 & 92.8 & 92.9 & 92.8 & 92.8 \\
\midrule
31 & 5000 & FaceFusion & Lip-Sync & 85 & 18 & 80.7 & 87.3 & 78.7 & 80.0 \\
32 & 5000 & Lip-Sync   & Lip-Sync & 66 & 14 & 97.1 & 97.5 & 97.1 & 97.5 \\
\bottomrule
\end{tabular}
\label{tab:performance-results}
\end{table*}

\subsection{Classification}

From the $9$D features extracted from each video (see Section~\ref{subsec:feature-engineering}), we employ the tree-based ensemble model XGBoost~\cite{chen2016xgboost} to classify a video as authentic or deepfake. There are, of course, many classification options and while any of a number of different approaches may be effective, we chose XGBoost for a few reasons: (1) boosting models are effective at capturing complex non-linear relationships between different classes~\cite{chen2016xgboost, nielsen2016tree}; (2) tree-based approaches have the benefit of providing feature importance metrics which is attractive for our situation where the extracted features are well-defined statistical measurements; and (3) XGBoost performs well on smaller datasets, and those with mixed-scale features such as ours~\cite{nielsen2016tree}.

XGBoost was trained to minimize the logistic loss function:
\begin{eqnarray}
\mathcal{L} & = & -\frac{1}{N}\sum_{i=1}^{N} -l_i \log(\hat{l}_i) - (1-l_i)\log(1-\hat{l}_i)
\end{eqnarray}
where the true label associated the $i^{th}$ video in the dataset is denoted as $l_i \in \{0, 1\}$ with a label of $l_i=0$ representing a real video, and $l_i=1$ representing a fake video, and the predicted label is denoted as $\hat{l}_i$. The predicted label is defined as $\hat{l}_i = \frac{1}{1 + \exp(-f(\vec{x}_i))}$, where $\vec{x}_i$ is the $10$D feature vector extracted from the $i^{th}$ video and $f(\cdot)$ is the raw output of the XGBoost model.

\begin{table*}[t]
\centering
\caption{Performance metrics  (Accuracy, Precision, Recall, and F1) across video resolutions (1-3), compression rates (4-6), and datasets (7-8); the final row 9 corresponds to training/evaluation on all data enumerated in rows 1-8 as well as all data in rows 1-32 of Table~\ref{tab:performance-results}.}
\vspace{0.15cm}
\small  % Uncomment this line
\begin{tabular}{c|cc|c|cc|cccc}
\toprule
& \textbf{Video} & \textbf{Train} & \textbf{Eval} & \textbf{Train} & \textbf{Eval} & \multicolumn{4}{c}{\textbf{Evaluation (\%)}} \\
& \textbf{Length} & \textbf{Model} & \textbf{Model} & \textbf{Size} & \textbf{Size} & \textbf{Acc} & \textbf{Prec} & \textbf{Recall} & \textbf{F1}\\
\midrule
1 & 5000 & FaceFusion & FaceFusion $\Downarrow 75\%$ resolution & 66 & 29 & 96.6 & 96.8 & 96.6 & 96.5 \\
2 & 5000 & FaceFusion & FaceFusion $\Downarrow 50\%$ resolution & 66 & 29 & 96.6 & 96.8 & 96.6 & 96.5 \\
3 & 5000 & FaceFusion & FaceFusion $\Downarrow 25\%$ resolution & 66 & 29 & 93.1 & 93.9 & 93.1 & 93.1 \\
\midrule
\midrule
4 & 5000 & FaceFusion & FaceFusion $\Downarrow 75\%$ bitrate & 66 & 37 & 71.6  & 79.4 & 71.6 & 69.6 \\
5 & 5000 & FaceFusion & FaceFusion $\Downarrow 50\%$ bitrate & 66 & 37 &  70.3 & 78.6 & 70.3 & 67.9 \\
6 & 5000 & FaceFusion & FaceFusion $\Downarrow 25\%$ bitrate & 66 & 37 & 66.2  & 70.6 & 66.2 & 64.3 \\
\midrule
\midrule
7 & 5000 & FaceFusion & Celeb-DF-v2  & 66 & 9 & 48.3 & 41.4 & 47.5 & 35.6 \\
8 & 5000 & Celeb-DF-v2 & Celeb-DF-v2  & 42 & 9 &  99.1 & 99.2 & 99.1 & 99.1 \\
\midrule
\midrule
9 & 5000 & All & All & 322 & 69 & 94.9 & 95.4 & 94.8 & 94.9 \\
\bottomrule
\end{tabular}
\label{tab:ablation}
\end{table*}

XGBoost incorporates a regularization term to control model complexity. We incorporate this regularization to constrain the model from using too many decision trees and from assigning extreme importance values to specific features. This regularization helps the model focus on robust patterns rather than memorizing the training data. This regularization takes the form:
\begin{eqnarray}
\cal{O} & = & \mathcal{L} + \gamma t + \lambda \sum_{t=1}^T |w_t|^2
\end{eqnarray}
where $O$ denotes the objective function to be optimized, $T=100$ represents the maximum number of trees, $w_t$ denotes the leaf weights in tree $t$, and $\gamma=0.1$ and $\lambda=1.0$ are the regularization parameters controlling tree complexity and L2 regularization, respectively.

XGBoost is further configured with a learning rate of 0.1, maximum tree depth of 6, a minimum child weight of 1, and subsampling and column sampling ratios of 0.8 each. To further prevent overfitting, we implemented early stopping with a patience of 10 rounds, monitoring validation loss (logloss) at each iteration. Training terminates automatically when validation loss fails to improve for 10 consecutive boosting iterations, indicating diminishing returns from additional trees. For hyperparameter optimization, we employed $k$-fold cross-validation ($k=5$) on the training set using GroupKFold to maintain identity separation, optimizing for weighted F1-score to account for potential class imbalance.

% ---------------------------------------------------------------------------------
\section{Results}

We report four metrics of performance: accuracy, precision, recall and F1-Score.  We denote a true positive (TP) as correctly classifying a deepfake face; a true negative (TN) as correctly classifying a natural face; a false positive (FP) as incorrectly classifying a natural face as a deepfake; and a false negative (FN) as incorrectly classifying a deepfake face as natural. Precision is then defined as TP/(TP + FP) -- this tells us the percentage of classified fakes are fake. Recall is TP/(TP+FN), which tells us the percentage of deepfakes caught. Accuracy is simply the average of the true positive (TP) and true negative (TN). The F1 score is defined as 2(Precision * Recall)/(Precision + Recall).
 
Shown in rows 1, 5, and 9 of Table~\ref{tab:performance-results} is the performance of our classifier when trained on $5000$ frames ($166$ seconds) from one type of face-swap deepfake (FaceFusion, FaceFusion+GAN, and FaceFusion Live) and evaluated on the same type of deepfake. In this matched condition where the training and evaluation were of a similar type, accuracy hovers between $94.7\%$ and $96.6\%$. In practice, of course, we will not know what type of deepfake we are confronted with and so generalizability across different training and evaluation is important.

Shown in rows 2, 3, 4, 6, 7, and 8 of Table~\ref{tab:performance-results} is the performance of the classifier when trained on $5000$ frames from one type of face-swap deepfake (FaceFusion, FaceFusion+GAN, and FaceFusion Live), but this time evaluated against each different type of face-swap deepfake. Here we see good generalization between FaceFusion and FaceFusion+GAN models with accuracy above $96\%$. However, generalization from these two models to FaceFusion Live, and vice versa, struggles with a $10$ percentage drop in performance. This led us to next investigate the efficacy of training a single model on all three face-swap deepfakes.

Shown in rows 10-12 are the results for a model trained on $5000$ frames of all three face-swap deepfakes (FaceFusion, FaceFusion+GAN, and FaceFusion Live). As before, accuracy for FaceFusion and FaceFusion+GAN is quite good at $98.3$\%, but FaceFusion Live continues to struggle achieving only $85.6$\% accuracy. As seen in row 9 of Table~\ref{tab:performance-results}, the FaceFusion Live videos can be distinguished from authentic video, but whatever artifacts that are present are distinct form the other FaceFusion models which our classifier is not able to generalize across.

So far our classifiers have been trained and evaluated on videos of a fixed length of $5000$ frames ($166$ seconds). Shown in Rows 13-15 of Table~\ref{tab:performance-results}. As seen in rows 13-15, longer videos provide no additional boost in accuracy. We next investigate how video length impacts classification accuracy.

Shown in rows 16-20 is the result of training and evaluating our classifier on FaceFusion videos of length $4000$, $3000$, $2000$, $1000$ and $500$ frames. The classifiers perform well up to $2000$ frames with accuracies between $96.2$\% and $94.7$\%, with a graceful reduction as video length decreases to $500$ frames.  The same general trend can be seen for FaceFusion+GAN (rows 21-25) and FaceFusion Live (rows 26-30).

Lastly, shown in rows 31 and 32 is the accuracy of our classifier when trained separately on $5000$ frames of face-swap and lip-sync deepfakes and evaluated against lip-sync deepfakes. When trained and evaluated on lip-sync deepfakes, performance is $97.1\%$, on par with the best face-swap accuracies described above. Somewhat surprisingly, when trained on face-swap and evaluated on lip-sync, performance only drops by $11$ percentage points. This is particularly surprising given that lip-sync deepfakes only modify the mouth region as compared to the entire face for face-swap deepfakes.

\subsection{Ablation}

Shown in rows 1-3 of Table~\ref{tab:ablation} are the results of training/evaluating on FaceFusion for videos at $75\%$, $50\%$, and $25\%$ of the original spatial resolution. As compared to an accuracy of $96.6\%$ at full resolution (row 1 of Table~\ref{tab:performance-results}), our classifier is robust across resolution, retaining an accuracy of $93.1\%$ accuracy even at a $25\%$ reduction in resolution.

By comparison, shown in rows 4-6 of Table~\ref{tab:ablation} is the impact of reducing the video bitrate to $75\%$, $50\%$, and $25\%$ of the original quality. Here accuracy immediately plummets to just $71.6\%$. This collapse in accuracy is due not to the inability to detect facial biometric differences but to a shift in the statistical nature of these differences (see Section~\ref{sec:methods}). In particular, a classifier retrained on representative samples across all bitrates yields an accuracy of $94.9\%$.

\begin{figure*}[t]
\begin{center}
\begin{tabular}{@{}ccccc@{}}
\includegraphics[width=0.18\textwidth]{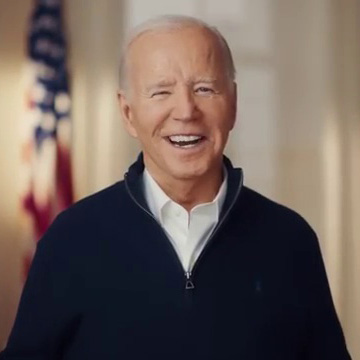} &
\includegraphics[width=0.18\textwidth]{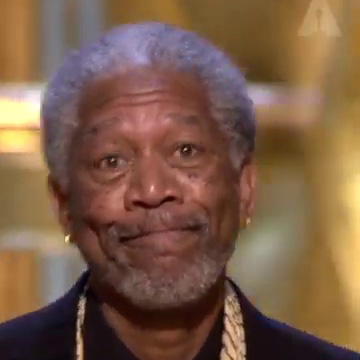} &
\includegraphics[width=0.18\textwidth]{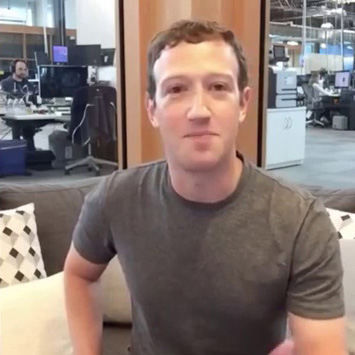} &
\includegraphics[width=0.18\textwidth]{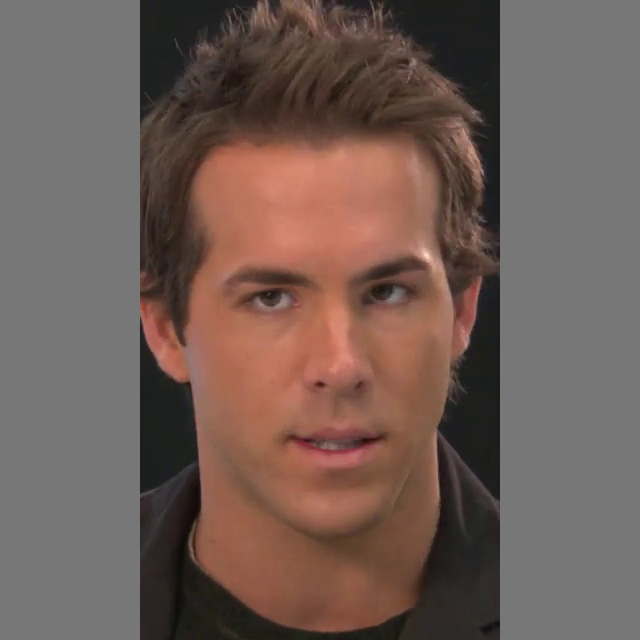} &
\includegraphics[width=0.18\textwidth]{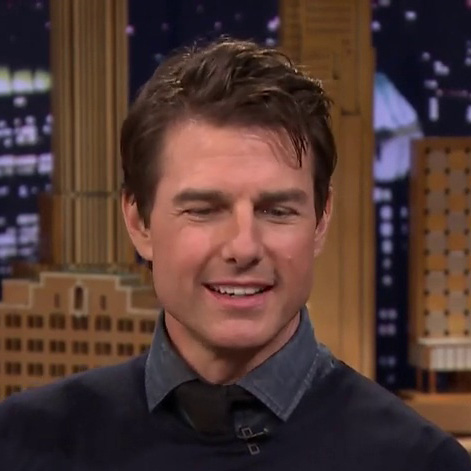} \\
\includegraphics[width=0.18\textwidth]{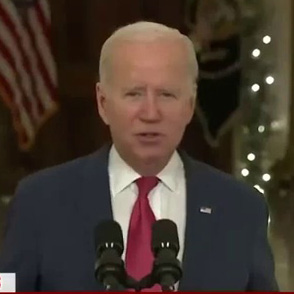} &
\includegraphics[width=0.18\textwidth]{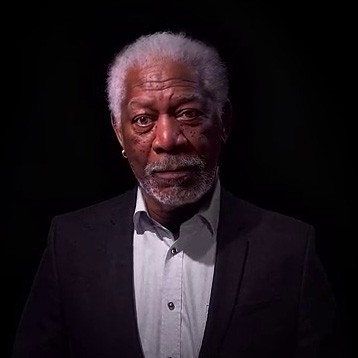} &
\includegraphics[width=0.18\textwidth]{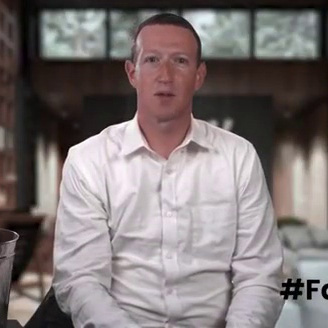} &
\includegraphics[width=0.18\textwidth]{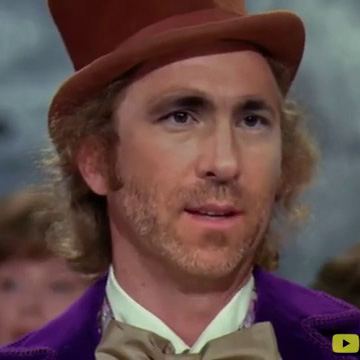} &
\includegraphics[width=0.18\textwidth]{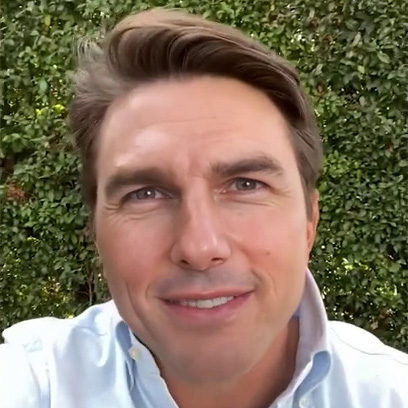}
\end{tabular}
\end{center}
\caption{Representative frames of real (top) and in-the-wild deepfake videos (bottom).}
\label{fig:in-the-wild}
\end{figure*}

Additionally, shown in rows 7-8 of Table~\ref{tab:ablation} is an evaluation of detection on the Celeb-DF-v2 dataset~\cite{li2019celeb, li2020celebdflargescalechallengingdataset} when trained on our FaceFusion dataset. With an accuracy of $48.3\%$, performance is again impacted to chance (row 7), but when retrained on data from Celeb-DF, accuracy rebounds to $99.1\%$ (row 8).

Finally, as shown in row 9 of Table~\ref{tab:ablation}, a classifier was trained on samples derived from all of the datasets: FaceFusion, FaceFusion Live, FaceFusion + GAN, LipSync, FaceFusion bitrate/resolution degradations, and Celeb-DF-V2. With an overall accuracy, precision, and recall of $94.9\%$ this model performs well across a diverse set of videos.

Generally speaking, our classifier shows good but not perfect generalizability across some video laundering and out-of-domain deepfakes. Augmenting the classifier with representative samples quickly improves efficacy. It remains to be seen how a broadly tuned classifier will respond to a continual stream of out-of-domain samples, or if a more powerful classifier will generalize more effectively. 

\subsection{In the Wild}

Shown in Figure~\ref{fig:in-the-wild} are examples of real (top row) and deepfakes (bottom row) sourced from YouTube. A total of ten videos, five real, five deepfakes, ranging in length from $36$ to $277$ seconds were classified with the ``all'' model in row 9 of Table~\ref{tab:ablation}. This model correctly classified nine of the ten videos and only mis-classified the deepfake of Mark Zuckerberg. Although a small evaluation, this analysis shows that our models can generalize to in-the-wild deepfakes.

\subsection{Feature Importance}

As described in Section~\ref{subsec:feature-engineering}, our models take as input nine summary statistics. We next evaluate the relative importance of each of these features and their combinations. With the ``all''  model in row 9 of Table~\ref{tab:ablation} we trained a total of 511 models for each of the combination of the $9$ features. Shown in Table~\ref{tab:feature-importance} are the subset of features that gave rise to the top 10 and bottom 10 performing models.

The best performing models ranged in feature size from $4$ to $7$. These top-performing models always incorporated one of the two variance ratios, and skewness appears in $9$ out of the $10$ best performing models. In contrast, the lowest performing models leveraged only $1$ or $2$ features, suggesting that no single feature is sufficient to distinguish between the real and deepfake faces. With only four features, the top-performing model yields a similar accuracy to the full model with all nine features (see row 9 of Table~\ref{tab:ablation}).

\begin{table*}[t]
\centering
\caption{Feature combinations ranked by cross-validation accuracy showing the top 10 and bottom 10 performing models.}
\vspace{0.15cm}
%\small
\begin{tabular}{r|c|c|l}
\toprule
\textbf{Rank} & \textbf{Acc (\%)} & \textbf{\# Features} & \multicolumn{1}{l}{\textbf{Features}} \\
\midrule
- & 94.9 & 9 & mean, median, variance, skew, kurtosis, q25, q75, var\_mean\_ratio, kurt\_var\_ratio \\
\midrule
\midrule
1 & 94.9 & 4 & skewness, kurtosis, q25, var\_mean\_ratio \\
2 & 94.7 & 5 & variance, skewness,  q75, var\_mean\_ratio, kurt\_var\_ratio \\
3 & 94.5 & 7 & mean, median, variance, skewness, kurtosis, q25, kurt\_var\_ratio \\
4 & 94.3 & 5 & variance, skewness, kurtosis,  q25, var\_mean\_ratio \\
5 & 94.1 & 6 & median, variance, skewness, q25, var\_mean\_ratio, kurt\_var\_ratio \\
6 & 94.0 & 7 & median, variance, skewness, kurtosis, q25, var\_mean\_ratio, kurt\_var\_ratio \\
7 & 93.9 & 5 & median, skewness, kurtosis, q25, var\_mean\_ratio \\
8 & 93.8 & 4 & kurtosis, q75, var\_mean\_ratio, kurt\_var\_ratio \\
9 & 93.7 & 4 & skewness, kurtosis, q25, kurt\_var\_ratio \\
10 & 93.6 & 4 & variance, skewness,  q25, kurt\_var\_ratio \\
\midrule
\midrule
502 & 63.9 & 1 & median \\
503 & 63.1 & 1 & variance \\
504 & 62.7 & 2 & q25, q75 \\
505 & 62.6 & 2 & mean, variance \\
506 & 62.1 & 2 & mean, q75 \\
507 & 61.7 & 1 & mean \\
508 & 61.6 & 1 & var\_mean\_ratio \\
509 & 61.5 & 1 & q25 \\
510 & 61.0 & 2 & median, q25 \\
511 & 57.6 & 1 & q75 \\
\bottomrule
\end{tabular}
\label{tab:feature-importance}
\end{table*}

The importance of kurtosis-related features aligns with the theoretical understanding of deepfake artifacts.  Kurtosis generally is a measure of the shape of a probability distribution, quantifying its tailedness.  As such, it can describe how thin or broad a distribution is. Deepfake face generation techniques, while producing visually convincing results on a frame-by-frame basis, struggle to maintain consistent high-dimensional embedding relationships that match the natural variance patterns of authentic videos. This manifests particularly in the tails and peaks of similarity distributions, precisely the aspects that kurtosis measures capture.

Kurtosis-variance-ratio measures the peakedness of a distribution normalized by its spread, effectively capturing how concentrated values are around the mean relative to the distribution's overall variability. The presence of this feature in $7$ out of $10$ the top performing models suggests that deepfake videos may exhibit distinctive tail behavior and concentration patterns in their similarity distributions that differ systematically from authentic videos.

The 25th quantile captures the boundary of the lowest quarter of values in the distribution. Examining Figure \ref{fig:pairwise-biometrics}, we see that the natural distributions (in blue) have very few values below $0.8$, while one of the deepfake distributions extends more substantially into the ranges $0.6$-$0.8$. We speculate this is caused by deepfake generation algorithms struggling to maintain consistent facial identity throughout the video particularly in the presence of large head motions. 

The variance-mean-ratio quantifies the variability of a distribution normalized by its central tendency. While both natural and deepfake distributions have similar means (around $0.9$), the distributions in Figure~\ref{fig:pairwise-biometrics} clearly show that deepfakes have a higher variance due to their bimodal characteristics. We speculate this bimodal distribution is caused by deepfake generation algorithms producing two distinct types of frames. First, high-quality synthetic frames that closely match the target identity (creating the peak around $0.9$). Secondly, lower-quality frames where the generation process struggles with complex poses, lighting, or expressions, (creating the secondary distribution around $0.7$-$0.8$).

Both mean and median appear in the top and bottom performing models. As is evident in Figure \ref{fig:pairwise-biometrics}, both natural and deepfake distributions center around similar values (approximately $0.9$). However, the underlying distribution shapes are dramatically different. The natural distributions show tight clustering, while deepfake distributions exhibit bimodality and a broader spread. These properties suggest that, while mean and median alone cannot distinguish between natural and deepfake videos due to their similar central tendencies, they become diagnostic when combined with shape-describing features that capture the distinctive distributional differences visible in the figure.

\subsection{Comparison to State of the Art}

In comparison to the state of the art deepfake detection on the Celeb-DF-V2 dataset, our proposed approach ranks second out of $24$ detection tools~\cite{NASKAR2024e25933,CHENG2024104263}. In comparison to our accuracy of $99.1\%$ (row 8 of Table~\ref{tab:ablation}), the top $24$ models range in accuracy from $65.6\%$ to $99.3\%$, with the top six models ranging from $96.85$ to $99.3\%$.

Though not the top score, our approach is attractive for its simplicity, efficiency, and ease of training. Most other approaches leverage deep neural networks or complex ensembles leading to computationally expensive training.  Our proposed approach however, is much more efficient leading to smaller, less complex models. As such it can easily be adapted for applications in real-time and at the edge.

% moved last table to Discussion for layout

% ---------------------------------------------------------------------------------

\section{Discussion}

Our approach based on measuring the distribution of biometric facial similarity has a few advantages: (1) it is effective in the face of a broad range of deepfake videos; (2) unlike many techniques that latch onto lower-level artifacts, we are more resilient across a large range of resolutions and compression qualities; and (3) unlike many techniques, we do not require a particularly large training dataset and the underlying computations are both easy to implement and computationally efficient.

Our approach, however, also has some disadvantages: (1) accuracy is at its best when the video in question is at least 60 seconds in length (2000 frames at 30 fps); (2) many face-swap and lip-sync deepfakes are based on manipulating a face in 2D; as these techniques extend into 3D, it is reasonable to assume that facial identity will be better preserved; and (3) although it would not be trivial, equipped with knowledge of this biometric anomaly, an adversary could implement a counter-attack that explicitly modifies the face on each frame to match a natural distribution.

Beyond the primary observation of how facial identify shifts over time in a deepfake impersonation video, we have taken a minimal approach to classifying these artifacts that leverages some simple statistical moments and a tree-based classifier. We are not committed to this particularly classification model and remain open to more sophisticated approaches that may be to achieve a few additional percentage point improvement in overall accuracy over the current $95\%$.

Although we have focused only on off-line video generation, it is possible that our approach would be effective in detecting real-time deepfakes used, for example, in impostor hiring scams~\cite{newman2022good}. Here, in fact, our technique may have a further advantage as the adversary needs to generate a deepfake at or near a frame rate of 30 fps making it more likely that we will see some inconsistencies in the facial appearance or geometry.

While we have focused on one category of deepfake videos -- face-swap and lip-sync -- we expect that this technique will also be effective against avatar deepfakes, but it remains to be seen how it holds up against the latest text-to-video engines (e.g.,~Sora, Veo, etc.).

As generative AI continues on its ballistic trajectory of creating increasingly more realistic images, voices, and videos, accurate, scalable, and robust detection will become increasingly more difficult and urgent. To this end, we are encouraged to see the recent deployment of robust imperceptible watermarks like SynthID~\cite{dathathri2024scalable} being, by default, added to all of Google's generative-AI products~\cite{pushmeet2025synthid}. Active approaches like this along with content credentials (\url{https://c2pa.org/post/contentcredentials}) are an essential part of a holistic solution that work alongside the type of passive approaches described here.

{
    \small
    \bibliographystyle{abbrv}
    \bibliography{main}
}

\end{document}